\documentclass[conference]{IEEEtran}
\IEEEoverridecommandlockouts

\usepackage{cite}
\usepackage{amsmath,amssymb,amsfonts}
\usepackage{algorithmic}
\usepackage{graphicx}
\usepackage{textcomp}
\usepackage{xcolor}
\usepackage{booktabs}
\usepackage{multirow}
\usepackage{hyperref}
\def\BibTeX{{\rm B\kern-.05em{\sc i\kern-.025em b}\kern-.08em
    T\kern-.1667em\lower.7ex\hbox{E}\kern-.125emX}}

\newcommand{\eg}{\emph{e.g.}}

\begin{document}

\title{Defer to Plan: Adaptive Multi-Agent Fusion for End-to-End V2X Driving\thanks{This work was supported by Key-Area Research and Development Program of Guangdong Province (2023B0909040001) and Shenzhen Science and Technology Program (KJZD20231023094701003).}}

\author{
\IEEEauthorblockN{Nuoran Li, Zhang Zhang, Yueran Zhao, Tianze Wang, Chao Sun}
\IEEEauthorblockA{\textit{Shenzhen Automotive Research Institute} \\
\textit{National Engineering Research Center of Electric Vehicles} \\
\textit{Beijing Institute of Technology} \\
Shenzhen, China \\
\{3220230398, 3120225185, zhaoyueran, 3120230483, chaosun\}@bit.edu.cn}
}

\maketitle

\begin{abstract}
Vehicle-to-everything-aided autonomous driving (V2X-AD) significantly enhances driving performance through information sharing. However, existing collaborative perception methods only optimize module-level perception capabilities and fail to effectively serve the ultimate planning and control tasks. We propose an end-to-end collaborative driving system that directly optimizes planning task performance. The system employs MotionNetwork to fuse historical temporal information, utilizes attention mechanisms to efficiently compress spatial features into compact tokens, and adaptively fuses multi-agent features through an autoregressive decoder. Additionally, we introduce Mixture-of-Experts (MoE) architecture to enhance the model's representation capacity for heterogeneous features. Experiments demonstrate that our method achieves a driving score of 79.72, surpassing the state-of-the-art CoDriving baseline (77.15) by 3.33\% in closed-loop evaluation while maintaining communication efficiency.
\end{abstract}


\section{Introduction}
\label{sec:intro}

\begin{figure}[h]
\vspace{-2mm}
\centering
\includegraphics[width=\linewidth]{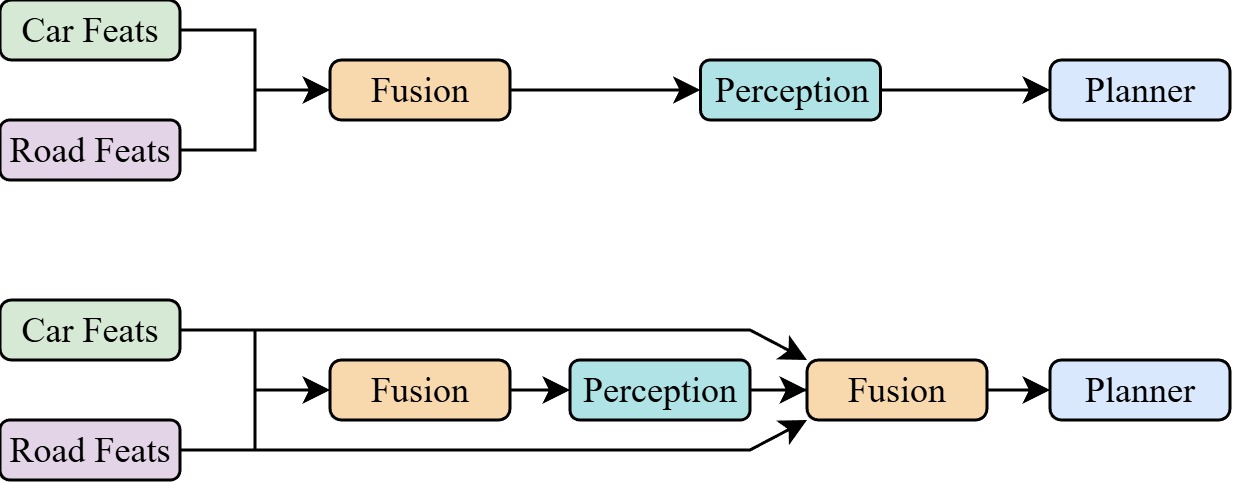}
\vspace{-3mm}
\caption{Perception-stage fusion vs Planning-stage fusion. Existing methods (top) perform multi-agent fusion at the perception stage with fixed weights. Our approach (bottom) defers fusion to the planning stage, enabling adaptive fusion through autoregressive decoder.}
\label{fig:paradigm}
\vspace{-3mm}
\end{figure}

Vehicle-to-everything-aided autonomous driving (V2X-AD) can significantly address the inherent limitations of single-vehicle autonomous driving, such as restricted field of view, occlusion blind spots, and unpredictability of other road users' behaviors. However, existing V2X collaborative driving methods uniformly adopt a \textbf{two-stage paradigm}: (1) multi-agent feature fusion is performed at the perception stage with fixed fusion weights; (2) the fused features are then passed to single-vehicle planning. This paradigm suffers from \textbf{fusion timing mismatch}---fusion strategies designed for perception tasks (\eg, fixed attention weights optimized for detection) may not be suitable for planning, where scenario-dependent adaptive weighting is required. Representative collaborative perception and prediction methods~\cite{v2vnet,v2xvit,v2xpnp} all perform fusion before the planning stage, leaving \textbf{no opportunity for planning-aware adaptive fusion}.

CoDriving~\cite{codriving}, the state-of-the-art end-to-end V2X system, exemplifies this limitation: multi-agent fusion occurs at the perception stage with fixed weights that cannot adapt to planning requirements (\eg, roadside features are critical in occluded intersections but less valuable on open highways). Additionally, roadside features contain 75\%--90\% background information redundant for planning.

\textbf{Our Core Insight:} The key issue is \textit{when} and \textit{how} to fuse multi-agent features. We propose to \textbf{defer fusion to the planning stage}, where an autoregressive decoder can dynamically learn adaptive fusion weights based on the current driving context, rather than using fixed weights determined at the perception stage.

\begin{figure}[t]
\vspace{-2mm}
\centering
\includegraphics[width=\linewidth]{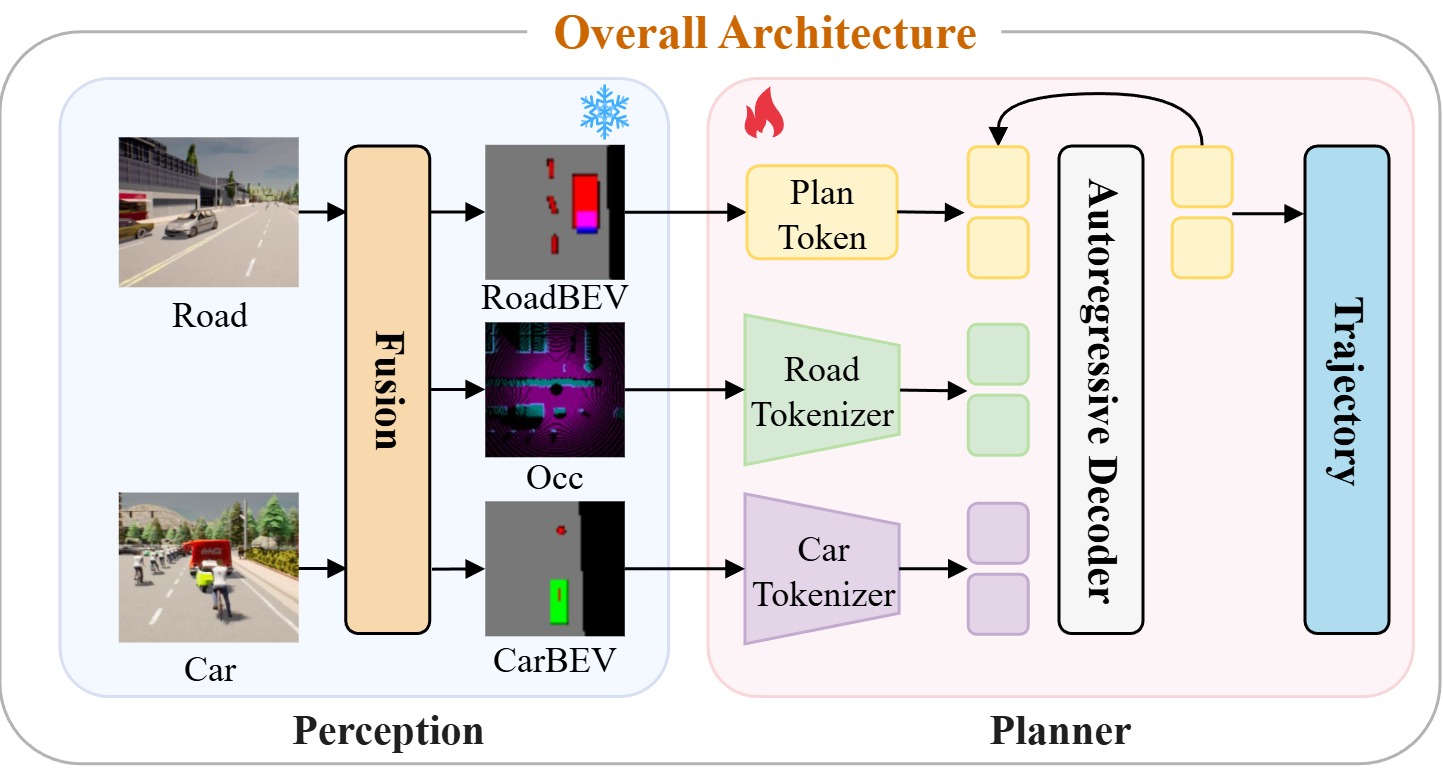}
\vspace{-3mm}
\caption{Overall architecture of our planning-oriented V2X collaborative driving system.}
\label{fig:overview}
\vspace{-2mm}
\end{figure}

\textbf{Our Approach.} As illustrated in Figure~\ref{fig:overview}, our system features:
\begin{itemize}
    \item \textbf{Heterogeneous Feature Processing:} Targeting feature differences between vehicle and infrastructure agents, we design a dual-stream processing pipeline with planning-oriented channel-wise purification on roadside features and MoE-enhanced tokenizers.
    \item \textbf{Autoregressive Adaptive Fusion:} Employ LLaMA-style autoregressive decoder to achieve adaptive fusion of multi-agent features, dynamically learning importance weights of ego and road tokens through self-attention mechanism.
    \item \textbf{End-to-End Planning Optimization:} Directly optimize planning tasks through end-to-end training, achieving complete pipeline optimization from feature extraction to trajectory generation.
\end{itemize}

\textbf{Contributions:}
\begin{itemize}
    \item We propose \textbf{a planning-stage fusion framework} that defers multi-agent feature fusion to the planning stage, enabling adaptive fusion weights learned through the autoregressive decoder.
    \item We design dual-stream feature processing with channel-wise purification and MoE-enhanced tokenization for heterogeneous vehicle-infrastructure features.
    \item We introduce autoregressive decoder that dynamically learns scenario-dependent fusion weights, replacing fixed fusion at the perception stage.
    \item Experiments on V2Xverse demonstrate significant improvements over both single-vehicle and perception-stage fusion baselines.
\end{itemize}

\section{Related Work}
\label{sec:related}

\textbf{Collaborative Perception and Prediction.}
V2X collaborative perception and prediction methods~\cite{v2vnet,v2xvit,zhao2025heatv2x,v2xpnp} mostly perform multi-agent fusion before planning, typically with fixed fusion strategies optimized for perception or forecasting. Roadside-specific methods~\cite{zhang2025height3d,zhang2025pillarid} further study infrastructure-view representations. These designs provide strong upstream perception, but leave limited room for planning-aware adaptive fusion.

\textbf{End-to-End Autonomous Driving.}
End-to-end driving methods~\cite{uniad,transfuser,tcp} demonstrate planning-oriented optimization, while UniV2X~\cite{univ2x} extends this direction to V2X cooperation. Related roadside representation and road-scene understanding works~\cite{zhang2025heightformer,zhang2025pillarmamba,wang2025roadformer,wang2025roadmamba,zhu2024lanemapnet} improve infrastructure-side modeling. Our work focuses on planning-stage adaptive fusion with robustness analysis under latency and noise.

\textbf{Collaborative Driving.}
Collaborative driving systems such as Coopernaut~\cite{coopernaut} and CoDriving~\cite{codriving} perform fusion at the perception stage with fixed weights, creating an information bottleneck before planning. Our work defers fusion to the planning stage with adaptive weights learned by autoregressive decoder.

\textbf{MoE and Autoregressive Models.}
MoE~\cite{shazeer2017} enables conditional computation with expert specialization; DriveMoE~\cite{drivemoe} introduces it to autonomous driving. LLaMA~\cite{llama} demonstrates autoregressive generation capabilities. We leverage both for adaptive multi-agent fusion.

\section{Method}
\label{sec:method}

\subsection{Problem Formulation}
Consider $N$ collaborative agents (vehicles + roadside units), given observations $\{X_i\}^N_{i=1}$, the goal is to maximize each vehicle's driving performance while satisfying communication bandwidth constraint $B$:
\begin{equation}
\max_{\theta,P} \sum_{i=1}^{N} d(\Phi_\theta(X_i, D_i, \{P_{j\rightarrow i}\}^N_{j=1})) \quad \text{s.t.} \quad \sum_{j\neq i} \|P_{j\rightarrow i}\| \leq B
\end{equation}
where $d(\cdot)$ is the driving performance metric, $\Phi_\theta$ is the end-to-end driving network, and $P_{j\rightarrow i}$ is the communication message from agent $j$ to $i$.

\subsection{Overall Architecture}
\begin{figure*}[t]
\vspace{-2mm}
\centering
\includegraphics[width=\textwidth]{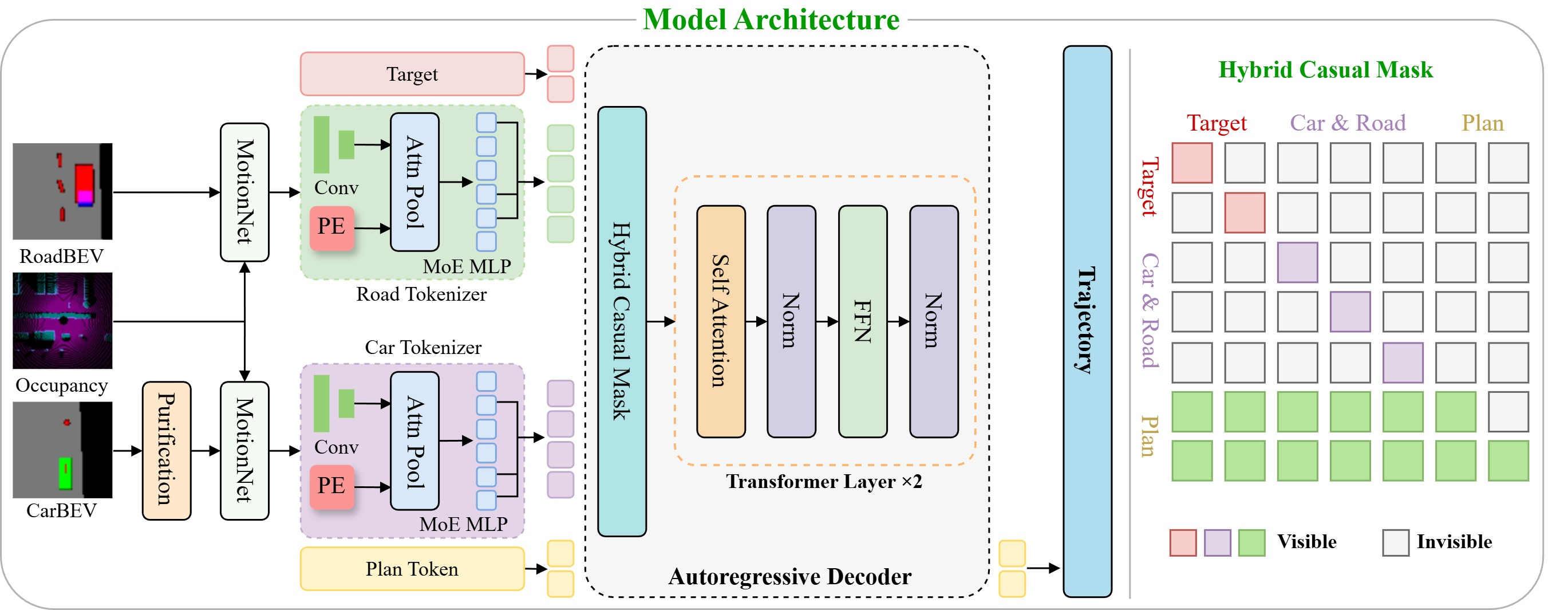}
\vspace{-3mm}
\caption{Detailed pipeline. Top: Ego stream processes BEV features through MotionNetwork and attention pooling. Bottom: Road stream applies channel-wise purification ($\alpha$=10\%) with MoE-enhanced tokenization (6 experts, Top-3 routing). Right: Autoregressive decoder with hybrid causal mask for adaptive fusion.}
\label{fig:architecture}
\vspace{-2mm}
\end{figure*}

As shown in Figure~\ref{fig:architecture}, our collaborative driving system adopts a dual-stream processing architecture. The pipeline consists of three main stages:

\textbf{Multi-Agent Feature Extraction:}
Vehicle and roadside agents separately collect LiDAR sensor data, extract spatial features through BEV encoder~\cite{lss,bevformer}, and generate occupancy maps.

\textbf{Dual-Stream Tokenization:}
The ego stream directly feeds BEV features through MotionNetwork for temporal aggregation, then applies attention-based tokenizer to extract compact ego tokens. The road stream first performs channel-wise purification to filter out planning-irrelevant background channels, then processes the filtered features through a separate MotionNetwork, and finally employs MoE-enhanced tokenizer to generate road tokens with expert specialization.

\textbf{Adaptive Multi-Agent Fusion:}
Construct token sequence and dynamically learn fusion weights through autoregressive decoder to generate collaborative waypoints.

\subsection{Multi-Agent Feature Extraction and Tokenization}

\subsubsection{Dual-Stream Temporal Feature Aggregation}
Inspired by CoDriving, we configure independent MotionNetworks for vehicle and roadside agents. Each MotionNetwork receives historical $T$ frames of BEV features $F^{(t)} \in \mathbb{R}^{T\times C\times H\times W}$ and occupancy maps $O^{(t)} \in \mathbb{R}^{T\times 6\times H\times W}$, performs spatiotemporal joint encoding through 3D convolution, and outputs compressed motion-aware features $M \in \mathbb{R}^{256\times H'\times W'}$.

\subsubsection{Planning-Oriented Road Feature Purification}
Roadside units have fixed viewpoints and wide coverage, with features often containing 75\%--90\% background information irrelevant to current planning tasks. We introduce channel-wise top-$k$ filtering before the roadside MotionNetwork:
\begin{enumerate}
    \item \textbf{Channel Activation:} $a_c = \text{mean}_{h,w}(F_c^{\text{road}}(h,w))$
    \item \textbf{Top-K Selection:} Retain top $\alpha\%$ channels ($\alpha=0.1$), zero out others
    \item \textbf{Temporal Fusion:} Filtered features enter MotionNetwork
\end{enumerate}
Ego vehicle features already focus on forward key regions and do not require purification.

\subsubsection{MoE-Enhanced Heterogeneous Tokenization}
We design FlexibleTokenizer to compress spatial features into a compact token set. A naive tokenizer with a single shared MLP struggles to simultaneously model ego-centric and infrastructure-centric views as well as diverse traffic patterns (highway, intersections, dense crowds). To address this, we instantiate independent tokenizers with separate parameters for ego and road, and further introduce MoE in the MLP layers so that different experts specialize on different driving maneuvers and scene patterns.

\textbf{Motivation for MoE Architecture.} Theoretically, MoE with $N$ experts offers two advantages over a single MLP: (1) \textit{Conditional computation}---activating only Top-$K$ experts reduces inference cost while maintaining model capacity; (2) \textit{Specialization}---different experts can specialize on distinct input patterns (\eg, sharp turns vs. straight driving), avoiding the mode-averaging effect where a single model learns a compromise representation suboptimal for all scenarios. However, these advantages only materialize if the router learns meaningful specialization; otherwise, the model degenerates to a weighted average of similar experts.

\textbf{Tokenization Pipeline.}
Features are projected and downsampled to $F' \in \mathbb{R}^{d\times H''\times W''}$, flattened with positional encoding: $X = \text{Flatten}(F') + \text{PE} \in \mathbb{R}^{H''W''\times d}$. Multi-head attention~\cite{attention} pooling with $k$ learnable queries $Q \in \mathbb{R}^{k\times d}$ aggregates spatial features:
\begin{equation}
\text{Tokens} = \text{MultiHeadAttention}(Q, X, X) \in \mathbb{R}^{k\times d}
\end{equation}

\textbf{MoE Integration with Trajectory-Guided Routing.}
We employ MoE layers with Top-3 routing among 6 experts. To prevent expert collapse, we introduce trajectory-guided supervision: K-means clustering on trajectory curvature identifies 6 maneuver patterns, providing soft labels for the router. The gating network computes $p = \text{Softmax}(\mathcal{R}(h))$, selecting Top-$K$ experts for token projection.

After processing, the outputs are $\text{ego\_tokens} \in \mathbb{R}^{B\times k_e\times d}$ and $\text{road\_tokens} \in \mathbb{R}^{B\times k_r\times d}$.

\subsection{Autoregressive Decoder for Multi-Agent Adaptive Fusion}

\subsubsection{From Fixed Fusion to Adaptive Collaboration}
CoDriving's fixed attention weights cannot adapt to scenarios---roadside features are critical in occluded intersections but less important on open roads. We propose autoregressive decoder for adaptive fusion, dynamically learning importance weights through self-attention.

\subsubsection{LLaMA-Style Parallel Autoregressive Decoder}
We adopt LLaMA-style~\cite{llama} parallel autoregressive decoder for efficient trajectory generation.

\textbf{Token Sequence Construction.}
To enable joint reasoning over navigation intent and multi-agent perception, we construct a unified token sequence by concatenating the target embedding, ego tokens, road tokens, and learnable planning queries:
\begin{equation}
S = [\text{target\_embed}; \text{ego\_tokens}; \text{road\_tokens}; \text{plan\_queries}]
\end{equation}
where $S \in \mathbb{R}^{(1+k_e+k_r+T_f)\times d}$, with $k_e$ and $k_r$ denoting the number of ego and road tokens respectively, and $T_f$ representing the number of future waypoints to predict.

\textbf{Hybrid Causal Mask.}
We design a hybrid attention mask that allows bidirectional attention within the context region (target + ego + road tokens) for comprehensive scene understanding, while enforcing causal attention in the planning region to ensure temporally coherent trajectory generation where each waypoint can only attend to previous waypoints and the full context. Concretely, the mask is applied to decoder self-attention after concatenating context tokens and planning queries into a single sequence. Context tokens attend to each other bidirectionally, while each planning query attends to the full context and only preceding planning queries. This preserves causal waypoint generation while retaining parallel decoding efficiency.

\textbf{Layer-wise Processing.}
The concatenated sequence is processed through multiple transformer decoder layers with the hybrid causal mask:
\begin{equation}
S^{(l+1)} = \text{TransformerLayer}(S^{(l)}, S^{(l)}, \text{mask}=M_{\text{causal}})
\end{equation}
This enables the planning queries to dynamically aggregate information from both ego and road tokens through attention weights learned end-to-end.

\textbf{Waypoint Generation.}
After transformer processing, the planning tokens are projected to 2D coordinates through a linear layer:
\begin{equation}
W = \text{Linear}(S_{\text{plan}}) \in \mathbb{R}^{T_f\times 2}
\end{equation}
Each row of $W$ represents the $(x, y)$ offset of a future waypoint relative to the ego vehicle's current position.

\subsubsection{Collaborative Information Flow}
In the decoder, multi-agent information flows through self-attention where ego and road tokens interact bidirectionally for mutual perception enhancement. Planning queries dynamically aggregate features from both streams through learned attention weights, adaptively emphasizing ego or road information based on driving context. The causal mask ensures sequential waypoint generation with temporal coherence.

\subsection{Training Strategy}
\textbf{Two-Stage Training.}
We first freeze the pre-trained BEV encoder and train planning components (MotionNetwork, tokenizers, decoder), then optionally fine-tune end-to-end with smaller learning rate.

\textbf{Loss Function.}
The training objective combines trajectory regression and MoE auxiliary losses:
\begin{equation}
\mathcal{L} = \mathcal{L}_{\text{waypoint}}(W_{\text{pred}}, W_{\text{gt}}) + \lambda_{\text{router}}(t) \mathcal{L}_{\text{router}} + \lambda_{\text{moe}} \mathcal{L}_{\text{balance}}
\end{equation}
where $\mathcal{L}_{\text{waypoint}}$ is L1 loss for trajectory regression.

\textbf{Preventing Expert and Router Collapse.} Without explicit guidance, MoE training faces two failure modes: (1) \textit{Expert collapse}---all experts learn similar representations, degenerating the model into a weighted average of redundant functions; (2) \textit{Router collapse}---the router consistently selects a fixed subset of experts, leaving others untrained. To prevent these failures, $\mathcal{L}_{\text{router}}$ provides trajectory-guided supervision from K-means clustering on curvature ($K$=6 maneuvers), and $\mathcal{L}_{\text{balance}}$ encourages balanced expert utilization. We apply curriculum annealing:
\begin{equation}
\lambda_{\text{router}}(t) = \lambda_0 \cdot \max(0, 1 - t/T_{\text{anneal}})
\end{equation}
where $\lambda_0$=0.03. This strategy applies strong supervision early to escape poor local minima, then progressively reduces guidance to allow autonomous expert selection.

\section{Experiments}
\label{sec:experiments}

\subsection{Experimental Setup}
We conduct experiments on V2Xverse~\cite{codriving}, a V2X simulation platform built on CARLA with closed-loop evaluation support. We evaluate on 67 routes across 8 towns with safety-critical scenarios including occluded intersections and unexpected pedestrians. We report closed-loop metrics including Driving Score (DS), Route Completion (RC), and Infraction Score (IS), where DS is the benchmark's overall driving score, RC measures route completion, and IS reflects infraction-related safety. We also report open-loop trajectory errors including Average Displacement Error (ADE) and Final Displacement Error (FDE), measured in meters. Baselines include single-agent methods (TCP~\cite{tcp}, TransFuser~\cite{transfuser}) and collaborative methods (CoDriving~\cite{codriving}, Coopernaut~\cite{coopernaut}, V2X-ViT~\cite{v2xvit}).

\begin{table}[h]
\centering
\caption{Single-GPU latency (trimmed mean, ms) under online inference (1$\times$ RTX 3090).}
\label{tab:latency}
\small
\begin{tabular}{lcc}
\toprule
Stage & Time (ms) & \% of total \\
\midrule
Perception & 158.2 & 89.6 \\
MotionNet & 5.9 & 3.3 \\
Tokenizer & 2.0 & 1.1 \\
AR Decoder & 10.2 & 5.8 \\
\textit{Planning total} & \textit{18.4} & \textit{10.4} \\
\midrule
Total & 176.6 & -- \\
\bottomrule
\end{tabular}
\vspace{-1em}
\end{table}

\subsection{Main Results}

Table~\ref{tab:main_results} summarizes both open-loop trajectory metrics (ADE/FDE) and closed-loop driving metrics (DS/RC/IS) on V2Xverse.

\begin{table}[h]
\centering
\caption{Main results on V2Xverse benchmark. ADE/FDE are open-loop trajectory errors; DS/RC/IS are closed-loop driving metrics.}
\label{tab:main_results}
\begin{tabular}{l|cc|ccc}
\toprule
Method & ADE$\downarrow$ & FDE$\downarrow$ & DS$\uparrow$ & RC$\uparrow$ & IS$\uparrow$ \\
\midrule
TCP~\cite{tcp} & -- & -- & 47.48 & 62.14 & 0.81 \\
Late Fusion & -- & -- & 52.40 & 90.72 & 0.57 \\
V2X-ViT~\cite{v2xvit} & 0.629 & 1.447 & 39.35 & 91.98 & 0.42 \\
Coopernaut~\cite{coopernaut} & 0.645 & 1.487 & -- & -- & -- \\
CoopDet3D~\cite{coopdet3d} & 0.623 & 1.439 & 63.04 & 88.12 & 0.71 \\
CoDriving~\cite{codriving} & 0.619 & 1.413 & 77.15 & \textbf{92.34} & 0.82 \\
\textbf{Ours} & \textbf{0.598} & \textbf{1.393} & \textbf{79.72} & 91.05 & \textbf{0.88} \\
\bottomrule
\end{tabular}
\end{table}

Our method achieves state-of-the-art performance across both evaluation paradigms. In open-loop evaluation, we surpass CoDriving by 3.39\% in ADE and 1.42\% in FDE, validating the effectiveness of our planning-oriented feature processing. In closed-loop driving, we achieve the highest driving score (79.72, +3.33\%) and infraction score (0.88, +7.32\%), demonstrating that our autoregressive adaptive fusion produces safer trajectories with fewer traffic violations. Compared with CoDriving, the slight RC decrease (-1.29\%) together with improved DS and IS suggests a safety-efficiency trade-off: our planner behaves more cautiously in some edge cases, sacrificing a small amount of completion for better infraction-related outcomes. The improvement in IS particularly validates our dual-stream design: channel-wise purification filters planning-irrelevant background from roadside features, while MoE tokenizers enable specialized processing for different driving maneuvers.
\subsection{Robustness Analysis}
We evaluate robustness under real-world communication challenges.

\textbf{Pose Estimation Noise.} Table~\ref{tab:noise_robustness} evaluates performance under Gaussian pose noise. Our method demonstrates superior robustness: under severe noise (0.6), we degrade by only 4.01\% ADE compared to CoDriving's 5.50\%. This stems from our autoregressive fusion, which dynamically down-weights misaligned roadside features.

\begin{table}[h]
\centering
\caption{Robustness to pose estimation noise (ADE/FDE in meters).}
\label{tab:noise_robustness}
\begin{tabular}{l|cc|cc}
\toprule
\multirow{2}{*}{Noise} & \multicolumn{2}{c|}{CoDriving} & \multicolumn{2}{c}{Ours} \\
& ADE$\downarrow$ & FDE$\downarrow$ & ADE$\downarrow$ & FDE$\downarrow$ \\
\midrule
0.0 & 0.618 & 1.412 & \textbf{0.599} & \textbf{1.405} \\
0.2 & 0.625 & 1.431 & \textbf{0.605} & \textbf{1.423} \\
0.4 & 0.644 & 1.477 & \textbf{0.618} & \textbf{1.452} \\
0.6 & 0.652 & 1.491 & \textbf{0.623} & \textbf{1.463} \\
\midrule
\shortstack{Performance Loss\\(0.6 vs 0.0)} & 5.50\% & 5.59\% & \textbf{4.01\%} & \textbf{4.13\%} \\
\bottomrule
\end{tabular}
\end{table}

\textbf{Communication Latency.} Network congestion can cause roadside information to arrive 200--600ms late. Table~\ref{tab:latency_robustness} shows our autoregressive fusion handles stale information gracefully: under 600ms latency, we degrade by only 2.34\% ADE vs. CoDriving's 2.75\%. The adaptive attention learns to rely more on ego features when roadside information becomes temporally misaligned.

\begin{table}[h]
\centering
\caption{Robustness to communication latency (ADE/FDE in meters).}
\label{tab:latency_robustness}
\begin{tabular}{l|cc|cc}
\toprule
\multirow{2}{*}{Latency} & \multicolumn{2}{c|}{CoDriving} & \multicolumn{2}{c}{Ours} \\
& ADE$\downarrow$ & FDE$\downarrow$ & ADE$\downarrow$ & FDE$\downarrow$ \\
\midrule
0ms & 0.618 & 1.412 & \textbf{0.599} & \textbf{1.405} \\
200ms & 0.620 & 1.419 & \textbf{0.599} & \textbf{1.400} \\
400ms & 0.626 & 1.435 & \textbf{0.600} & \textbf{1.402} \\
600ms & 0.635 & 1.453 & \textbf{0.613} & \textbf{1.425} \\
\midrule
\shortstack{Performance Loss\\(600ms vs 0ms)} & 2.75\% & 2.90\% & \textbf{2.34\%} & \textbf{1.42\%} \\
\bottomrule
\end{tabular}
\end{table}

These results validate that our autoregressive fusion dynamically re-weights ego and roadside tokens to compensate for degraded collaborative information.

\subsection{Ablation Studies}

To validate the effectiveness of each component, we conduct ablation studies. Table~\ref{tab:ablation} shows the results.

\begin{table}[h]
\centering
\caption{Ablation study on key components.}
\label{tab:ablation}
\begin{tabular}{l|cccc}
        \toprule
Configuration & ADE$\downarrow$ & FDE$\downarrow$  \\
        \midrule
Full Model (Ours) & \textbf{0.598} & \textbf{1.393}  \\
w/o MoE Tokenizers & 0.610 & 1.406 \\
w/o Channel-wise Purification & 0.612 & 1.412  \\
w/o Autoregressive Decoder & 0.628 & 1.447  \\
CoDriving Baseline & 0.619 & 1.413  \\
        \bottomrule
    \end{tabular}
\end{table}

Our method builds on CoDriving with three key components:
\begin{itemize}
    \item \textbf{MoE-Enhanced Tokenizers:} Specialized experts yield 1.84\% ADE and 0.91\% FDE gains through conditional expert selection for different driving maneuvers.
    \item \textbf{Channel-wise Purification:} Roadside feature purification contributes 2.34\% ADE and 1.36\% FDE improvements by suppressing planning-irrelevant background channels.
    \item \textbf{Autoregressive Decoder:} Replacing fixed fusion brings the largest gain (4.85\% ADE, 3.88\% FDE) via dynamic re-weighting between ego and roadside tokens.
\end{itemize}
Together, these components achieve 4.21\% ADE and 3.28\% FDE improvement over CoDriving.

\subsection{Visualization and Analysis}

\subsubsection{Trajectory Visualization}

\begin{figure}[h]
\vspace{-2mm}
\centering
\includegraphics[width=\linewidth]{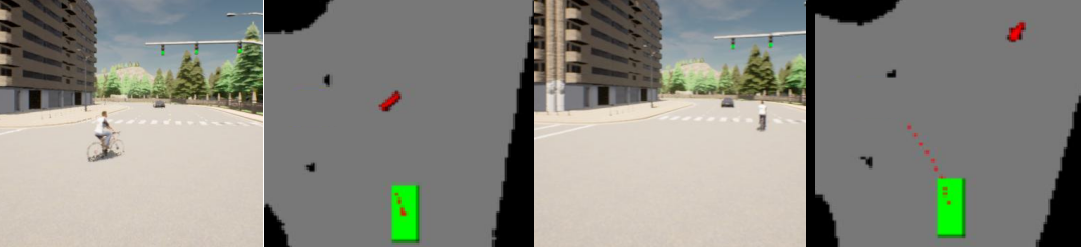}
\vspace{-3mm}
\caption{Trajectory visualization in a yield-and-turn scenario. When a cyclist suddenly appears during a left turn, our method stops to yield, then resumes after the cyclist passes.}
\label{fig:visualization}
\vspace{-3mm}
\end{figure}

Figure~\ref{fig:visualization} illustrates a challenging yield-and-turn scenario. When a cyclist emerges from an occluded region, our method: (1) promptly stops to yield, leveraging roadside features that detect the cyclist earlier; (2) smoothly resumes after the cyclist passes. This validates that our fusion dynamically adjusts ego vs. roadside importance based on the evolving context.

\subsubsection{MoE Expert Activation Patterns}

\begin{figure}[h]
\vspace{-2mm}
\centering
\includegraphics[width=\linewidth]{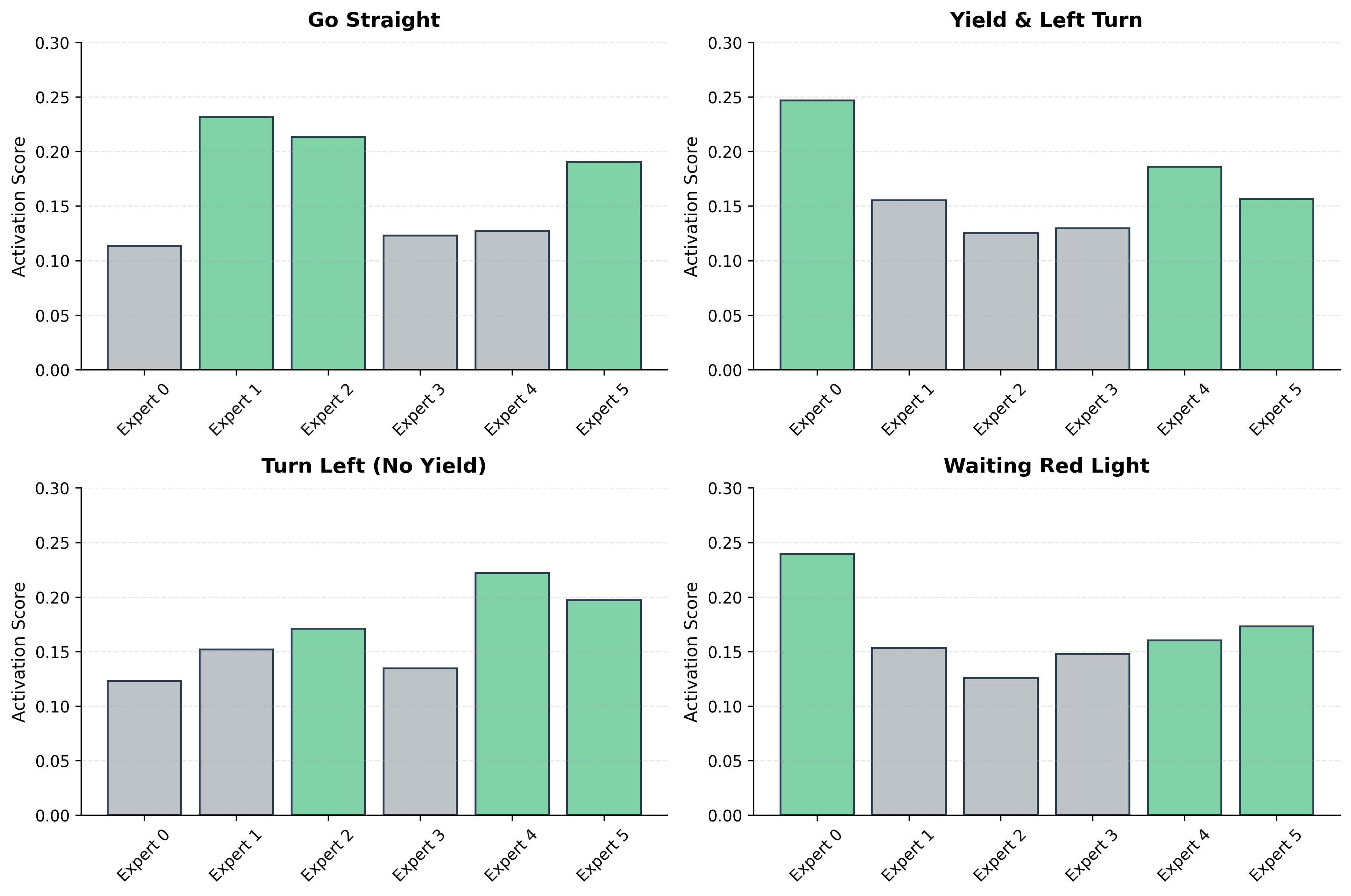}
\vspace{-3mm}
\caption{MoE expert activation patterns across four scenarios. Different maneuvers activate distinct expert combinations (Top-3 marked with \checkmark), validating that trajectory-guided routing prevents expert collapse.}
\label{fig:moe}
\vspace{-3mm}
\end{figure}

Figure~\ref{fig:moe} reveals meaningful expert specialization: straight driving activates Experts 1, 2, 5, while yielding with left turns triggers Experts 0, 4, 1. This confirms our trajectory-guided routing prevents expert collapse---each expert specializes on distinct maneuvers rather than degenerating into redundant copies.

\subsubsection{Feature Purification Effect}

\begin{figure}[h]
\vspace{-2mm}
\centering
\includegraphics[width=0.8\linewidth, angle=0]{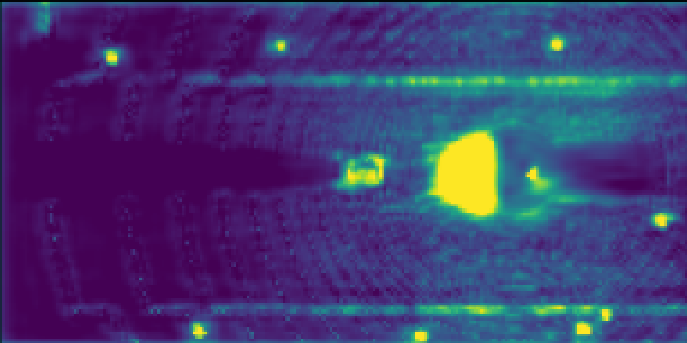}
\vspace{-3mm}
\caption{Channel-wise purification effect on roadside BEV features. The heatmap shows feature activation after Top-10\% channel selection, demonstrating focused attention on planning-critical regions.}
\label{fig:purification}
\vspace{-3mm}
\end{figure}

Figure~\ref{fig:purification} shows that after Top-10\% channel selection, feature activation concentrates on planning-critical regions: driving corridor, dynamic agents, and conflict zones. Background regions (distant buildings, parked vehicles) are suppressed, explaining the 2.34\% ADE improvement in our ablation study.

\section{Conclusion}
\label{sec:conclusion}

We propose a planning-oriented V2X collaborative driving system that defers multi-agent fusion to the planning stage, enabling adaptive weights learned through an autoregressive decoder. Our dual-stream pipeline with channel-wise purification and MoE-enhanced tokenizers effectively handles heterogeneous vehicle-infrastructure features. Experiments demonstrate state-of-the-art performance with 3.33\% driving score and 7.32\% infraction score improvements over CoDriving, along with stronger robustness under noise and latency.

\textbf{Limitations and Future Work.} Our system assumes fixed communication topology; future work can explore dynamic agent selection based on relevance. Explicitly modeling communication delays and incorporating uncertainty estimation would further enhance real-world applicability.

\vspace{-2mm}
\bibliographystyle{IEEEbib}
\bibliography{references}

\end{document}